\documentclass[letterpaper]{article}
\usepackage{aaai18}
\usepackage{times}
\usepackage{helvet}
\usepackage{courier}
\usepackage{url}
\usepackage{graphicx}
\usepackage{latexsym}
\usepackage{amsmath}
\usepackage{multirow}
\usepackage{booktabs}
\usepackage{array}
\usepackage{subfigure}
\usepackage{amssymb}

\title{Does William Shakespeare REALLY Write Hamlet? Knowledge Representation Learning with Confidence}
\author{Ruobing Xie$^{1,2}$, Zhiyuan Liu$^{1}$\thanks{Corresponding author: Z. Liu (liuzy@tsinghua.edu.cn)}, Fen Lin$^{2}$, Leyu Lin$^{2}$\\
$^{1}$ Department of Computer Science and Technology, \\
State Key Lab on Intelligent Technology and Systems, \\
National Lab for Information Science and Technology, Tsinghua University, China \\
$^{2}$ Search Product Center, WeChat Search Application Department, Tencent, China.
}

\date{}

\begin{document}

\maketitle

\begin{abstract}
  Knowledge graphs (KGs), which could provide essential relational information between entities, have been widely utilized in various knowledge-driven applications. Since the overall human knowledge is innumerable that still grows explosively and changes frequently, knowledge construction and update inevitably involve automatic mechanisms with less human supervision, which usually bring in plenty of noises and conflicts to KGs. However, most conventional knowledge representation learning methods assume that all triple facts in existing KGs share the same significance without any noises. To address this problem, we propose a novel confidence-aware knowledge representation learning framework (CKRL), which detects possible noises in KGs while learning knowledge representations with confidence simultaneously. Specifically, we introduce the triple confidence to conventional translation-based methods for knowledge representation learning. To make triple confidence more flexible and universal, we only utilize the internal structural information in KGs, and propose three kinds of triple confidences considering both local and global structural information. In experiments, We evaluate our models on knowledge graph noise detection, knowledge graph completion and triple classification. Experimental results demonstrate that our confidence-aware models achieve significant and consistent improvements on all tasks, which confirms the capability of CKRL modeling confidence with structural information in both KG noise detection and knowledge representation learning.
\end{abstract}

\section{Introduction}

Recent years have witnessed the great thrive in artificial intelligence that has broad impacts on our daily lives. In tasks like information retrieval and question answering, people are not satisfied with merely semantic matching, but expect AI agents to have knowledge for understanding, reasoning and solving. Knowledge graphs (KGs), which provide effective well-structured relational information between entities, are essential supporters for knowledge-based AI agents. A typical KG usually stores knowledge with triple facts in the form of (\emph{head entity}, \texttt{relation}, \emph{tail entity}), which is also abridged as $(h,r,t)$.

There are existing amounts of widely-utilized large-scale knowledge graphs such as Freebase \cite{bollacker2008freebase}, DBpedia \cite{auer2007dbpedia} and other domain-specific KGs. However, these knowledge graphs are still far from complete to describe the infinite real-world facts, in which some kinds of knowledge may even change frequently. Therefore, knowledge construction and timely update are significant for knowledge-driven applications. Most conventional knowledge graph construction methods usually involve huge human supervision or expert annotation, which are extremely labor-intensive and time-consuming. Nowadays, automatic mechanism and crowdsourcing take larger parts in knowledge construction, while these methods may suffer from possible noises and conflicts due to limited human supervision. For instance, recent neural relation extraction model on benchmark achieves only around $60\%$ precision when the recall is $20\%$ \cite{lin2016neural}. Moreover, \cite{heindorf2016vandalism} focuses on vandalism detection in Wikidata, which also verifies the existence and problems of noises in KGs.

In this paper, we concentrate on how to deal with noises in knowledge representation learning (KRL), which provides an effective and flexible way for using knowledge. KRL represents entities and relations with distributed representations mainly according to triple facts in KGs. Therefore, it is crucial to consider noises in knowledge representation learning and knowledge-driven tasks.

\begin{figure}[!htbp]
\centering
\includegraphics[width=0.99\columnwidth]{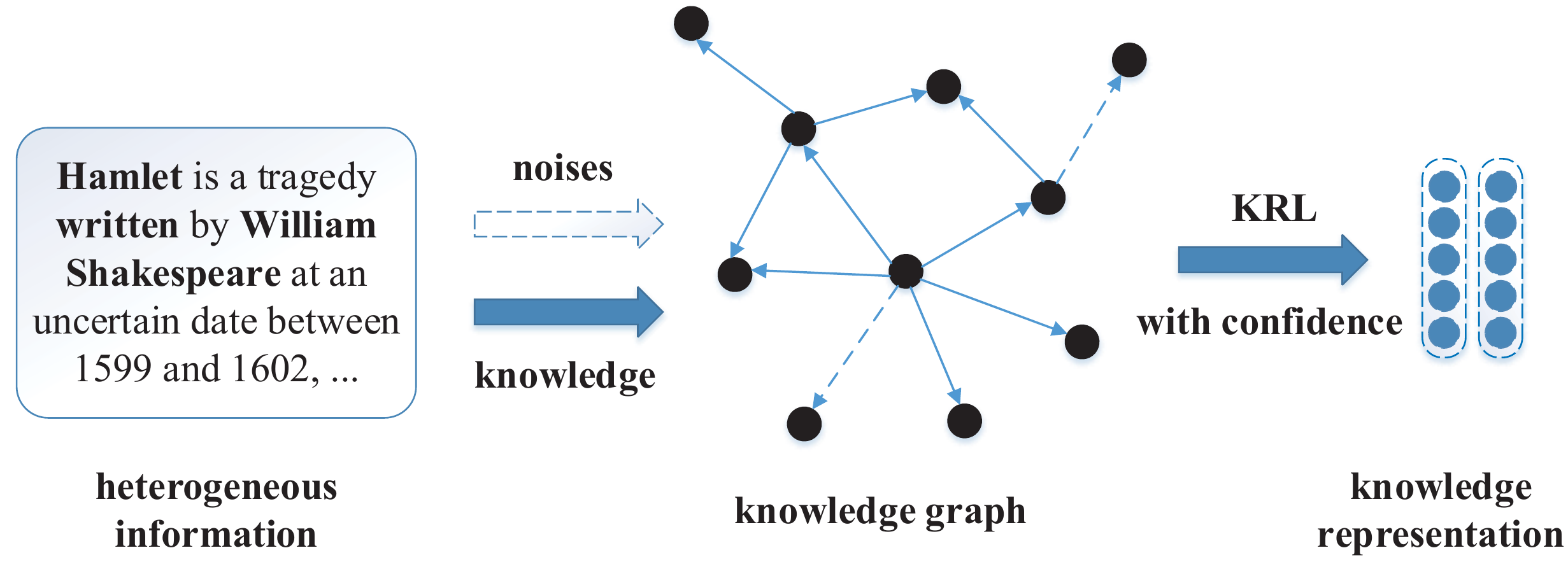}
\caption{Noises in KGs and confidence-aware KRL.}\label{fig. 1}
\end{figure}

We attempt to detect possible noises and conflicts located in existing knowledge graphs, while constructing noise-free knowledge representations simultaneously. However, most conventional KRL methods assume that all triple facts in existing KGs are completely correct. To solve this problem, we propose a novel confidence-aware knowledge representation learning (CKRL) framework taking possible noises into consideration. The notions of confidence and trust have been widely studied in fields such as cognitive science. Fig. \ref{fig. 1} demonstrates a brief illustration of our confidence-aware KRL framework, where the knowledge graph has both knowledge and noises extracted from heterogenous sources. These noises are expected to be detected via our confidence-aware model and to be ignored in knowledge representation learning.

Specifically, CKRL follows the translation-based framework proposed by \cite{bordes2013translating}, and learns knowledge representations with triple confidences. We propose three triple confidences considering both local triple and global path information, and knowledge representations are learned to fit the global consistency under translation assumption weighted by those dynamic triple confidences. To make the triple confidence more universal and flexible, we only consider the internal structural information in KGs, which correspondingly makes noise detection much more challenging due to the limited information.

In experiments, we evaluate our models on three tasks including knowledge graph noise detection, knowledge graph completion and triple classification. The results demonstrate that our models achieve the best performances on all tasks, which confirm the capability of CKRL in noise detection and knowledge representation learning. The main contributions of this work are concluded as follows:
\begin{itemize}
  \item We propose a novel confidence-aware KRL framework for knowledge graph noise detection and knowledge representation learning simultaneously, which only uses internal structural information in KGs.
  \item We evaluate our CKRL models on several datasets with different noise rates extended from a real-world dataset, and achieve promising performances on all tasks.
  \item The idea of triple confidence in CKRL could be utilized not only in knowledge representation learning, but also in knowledge construction.
\end{itemize}

\section{Related Work}

\subsection{Knowledge Graph Noise Detection}

It seems to be inevitable that noises do exist in KGs, which can strongly affect knowledge acquisition \cite{manago1987noise}. Moreover, a novel task named Wikidata vandalism, which aims to combat with deliberate destructions in knowledge graphs, has attracted wide attention \cite{heindorf2015towards}. Therefore, noise detection is essential in knowledge construction and knowledge application. Most knowledge graph noise detection works happen when constructing knowledge graphs. For instance, YAGO2 extracts knowledge from Wikipedia with human supervision that human judges are presented with selected facts for which they have to assess the correctness \cite{hoffart2013yago2}. Wikidata also relies on a crowd-sourced human curation software in which contributors can reject or approve a statement \cite{pellissier2016freebase}. DBpedia creates its mappings to Wikipedia infoboxes via a worldwide crowd-sourcing effort \cite{lehmann2015dbpedia}. These noise detections in large-scale KGs are usually involved with huge human efforts, which are extremely labor-intensive and time-consuming.

Recently, there are also lots of researches focusing on automatic KG noise detection \cite{nickel2016review}. However, most existing methods mainly concentrate on feature selection from contents, users, items and revisions \cite{heindorf2016vandalism}, and thus are constrained by the completeness of external information. There are also some efforts working on judging importance in graphs for nodes \cite{gyongyi2004combating} or for edges \cite{de2012novel}, but few works concentrating on the confidence of each triple. Knowledge Vault \cite{dong2014knowledge} proposes a joint approach with both web content (e.g., texts, tabular data and human annotations) and prior knowledge derived from existing KGs to judge triple qualities, whose performance strongly depends on the external information. In this paper, we introduce three triple confidences to KG noise detection and knowledge representation learning, which only focus on the internal structural information in KGs.

\subsection{Translation-based KRL Methods}

Recent years many efforts concentrate on learning distributed representations for knowledge graphs, among which the translation-based methods are both straightforward and effective with the state-of-the-art performances. TransE \cite{bordes2013translating} projects both entities and relations into a continuous low-dimensional vector space, interpreting relations as translating operations between head and tail entities. The translation assumption in TransE implies the equation that $\mathbf{h}+\mathbf{r} \simeq \mathbf{t}$. The energy function is defined as follows:
\begin{equation}
\begin{split}
E(h,r,t)=||\mathbf{h}+\mathbf{r}-\mathbf{t}||.
\end{split}
\end{equation}
TransE can well balance both effectiveness and efficiency compared to traditional methods, while the over-simplified translation assumption constrains the performance when dealing with complicated relations. Some enhanced KRL methods based on TransE attempt to solve this problem with translations on relation-specific hyperplanes \cite{wang2014transH}, relation-specific entity projection \cite{lin2015learning} and type-specific entity projection \cite{xie2016representation_t}. Moreover, the translation assumption only focuses on the local information in triples, which may fail to make full use of global graph information in KGs. \cite{lin2015modeling} extends TransE by encoding multi-step relation path information into knowledge representation learning. However, most conventional KRL methods assume that all triples in KG share the same confidence, which is inappropriate especially for those KGs constructed automatically with less human supervision. To the best of our knowledge, our model is the first embedding method to consider triple confidences of existing KGs in KRL. In this paper, we extend TransE to learn knowledge representations from noisy KGs, and it is not difficult for other enhanced translation-based methods to utilize our confidence-aware KRL framework.

\section{Methodology}

\begin{figure*}[!htbp]
\centering
\subfigure[Local triple confidence]{
\label{Fig.sub.2.1}
\includegraphics[width=0.48\textwidth]{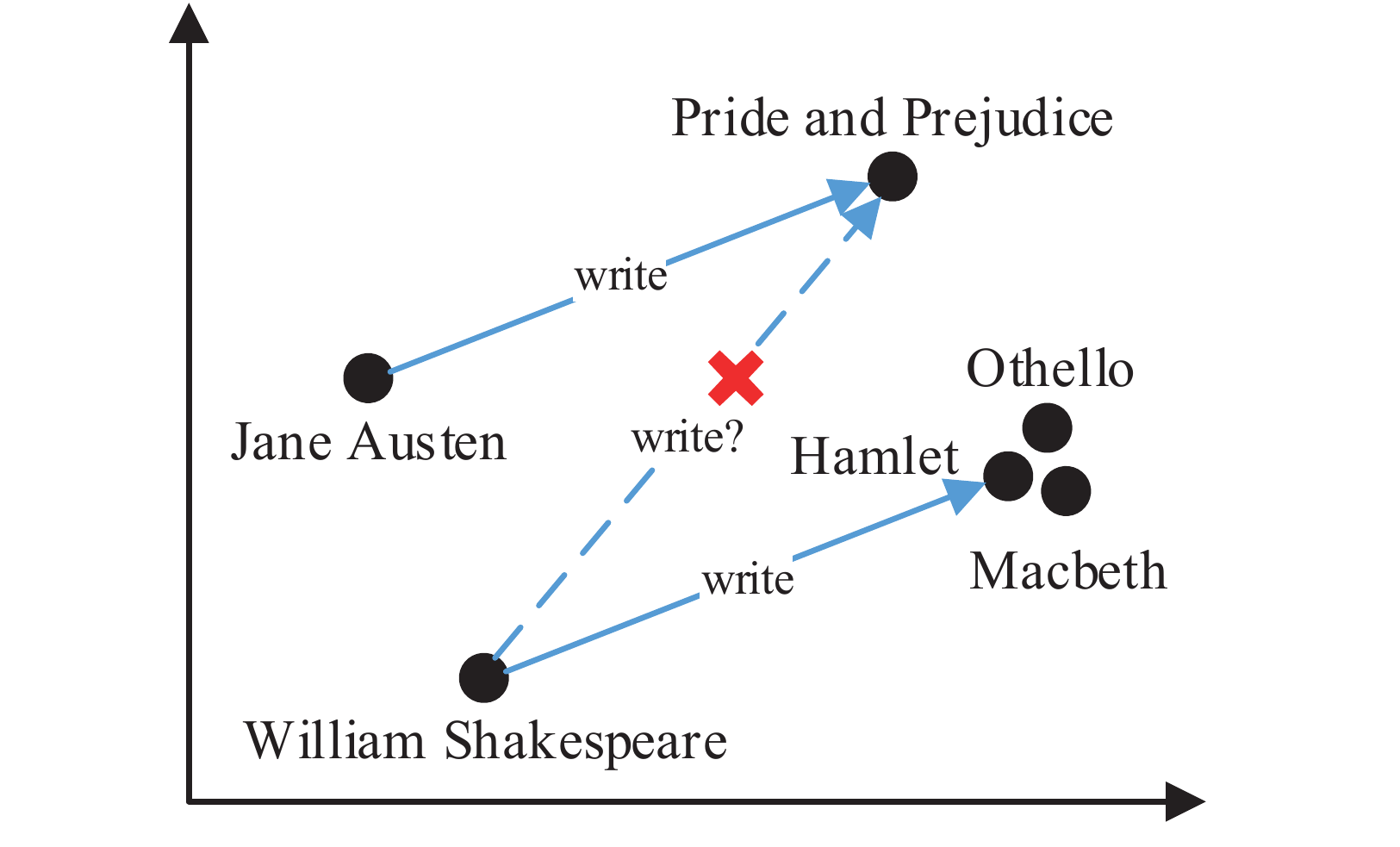}}
\subfigure[Global path confidence]{
\label{Fig.sub.2.2}
\includegraphics[width=0.48\textwidth]{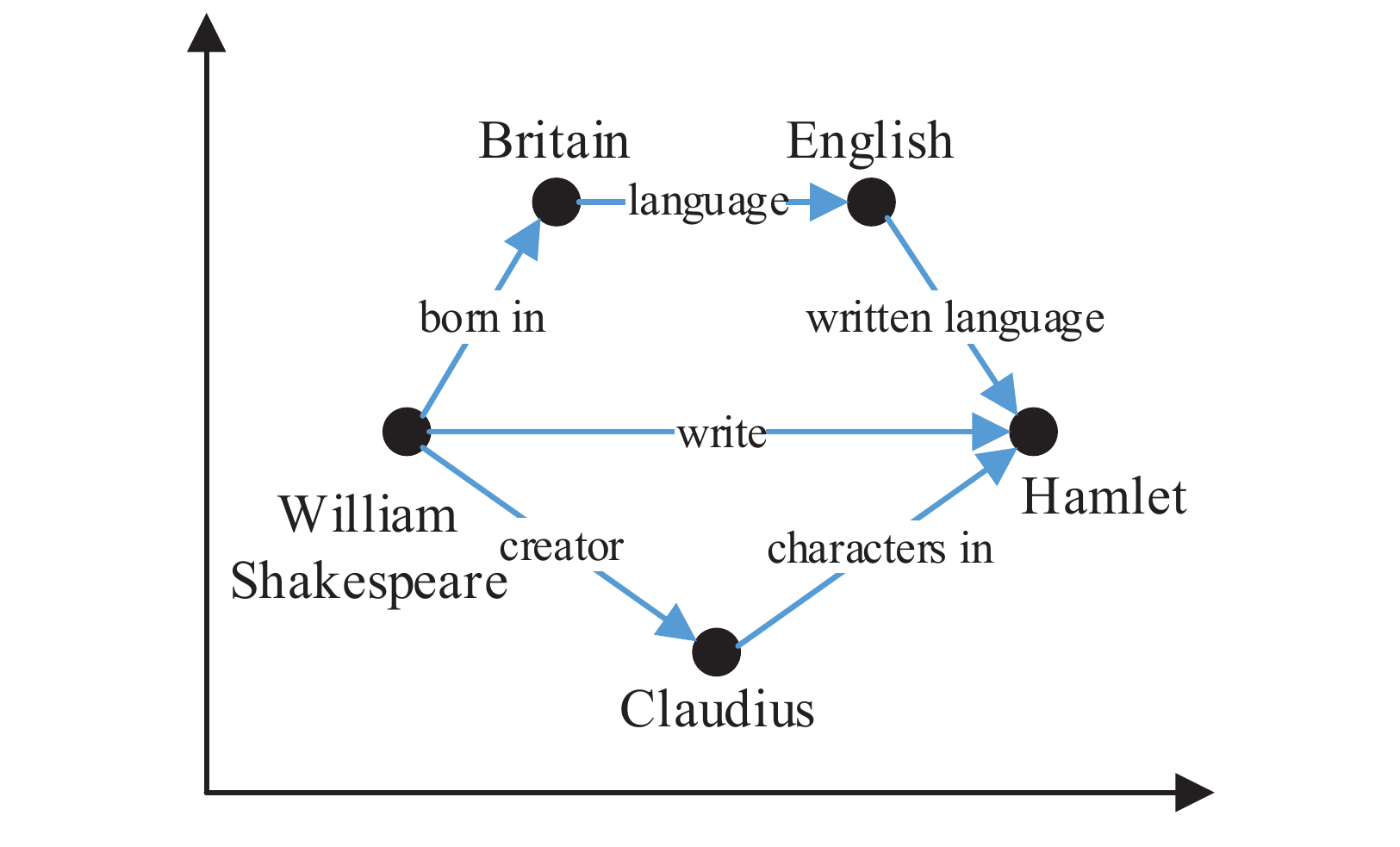}}
\caption{Effective mechanism of local triple confidence and global path confidence.}
\label{fig. 2}
\end{figure*}

We first give the notations used in this paper. Given a triple fact $(h,r,t)$, we consider the head and tail entities $h,t \in E$ and the relation $r \in R$, where $E$ and $R$ stand for the sets of entities and relations. $T$ represents the overall training triple facts including possible conflicts and noises.

To detect possible noises in knowledge graphs and learn better knowledge representations, we introduce a novel concept \textbf{triple confidence} for each triple fact. Triple confidence describes the correctness and significance of a triple, which could be measured with the favor of both internal structural information and external heterogeneous information.

\subsection{Confidence-aware KRL Framework}

We attempt to detect noises and learn better knowledge representations with triple confidence taken into consideration, concentrating more on those triples with high confidences. Following the translation-based framework, we design our confidence-aware KRL energy function as follows:
\begin{equation}
\begin{split}
E(T)=\sum_{(h,r,t)\in T}E(h,r,t) \cdot C(h,r,t).
\end{split}
\end{equation}
The confidence-aware energy function can be divided into two parts: $E(h,r,t)=||\mathbf{h}+\mathbf{r}-\mathbf{t}||$ stands for the dissimilarity score under translation assumption, which is the same as that of TransE. A lower $E(h,r,t)$ indicates that the entity and relation representations of this triple fit better with our translation framework. While differing from conventional methods, we also introduce the triple confidence $C(h,r,t)$ as the second part of our energy function. A higher triple confidence implies that the corresponding triple is more credible, and thus should be more considered.

Triple confidence can be calculated both during and after knowledge graph construction, from various aspects including internal information such as graphic evidence and external information such as textual evidence. To make our triple confidence more universal and practical, we only consider the internal structural information after KG construction in our model, and propose both local and global triple confidences which are iteratively optimized during training. In CKRL, we bring in the confidences of different triples to learn more about those significant triples, and thus could get better knowledge representations.

\subsection{Objective Formalization}

We introduce the detailed training objective of our model in this section. Following TransE \cite{bordes2013translating}, we formalize a margin-based score function with negative sampling as objective for training. This pair-wise score function attempts to make the scores of positive triples to be higher than those of negative triples. We have:
\begin{equation}
\begin{split}
L=\sum_{(h,r,t)\in T}\sum_{(h',r',t')\in T'}\max(0,\gamma+E(h,r,t)\\
-E(h',r',t')) \cdot C(h,r,t),
\end{split}
\end{equation}
where $E(h,r,t)$ is the dissimilarity score of positive triple and $E(h',r',t')$ is that of negative triple. $\gamma > 0$ is the hyper-parameter of margin, and $T'$ represents the negative triple set. Here the triple confidence $C(h,r,t)$ instructs our model to pay more attention on those more convincing facts.

For pair-wise training, since there are no explicit negative triples in knowledge graphs, we sample negative triples complying with the following rules:
\begin{equation}
\begin{split}
T'=&\{(h',r,t)|h'\in E\}\cup\{(h,r,t')|t'\in E\}\\
&\cup\{(h,r',t)|r'\in R\}, \quad (h,r,t)\in T.
\end{split}
\end{equation}
It means that one entity or relation in a positive triple is randomly replaced by another entity or relation in the overall set. Note that differing from TransE, we also add relation replacements for better performances in relation prediction. We also discard all triples already in $T$ from $T'$ to make sure our generated negative triples are truly negative.

\subsection{Local Triple Confidence}

We first come up with the local triple confidence (LT) which only concentrates on the inside of a triple. Since our CKRL framework follows the translation assumption that $\mathbf{h}+\mathbf{r} \simeq \mathbf{t}$, it is straightforward to directly utilize this dissimilarity function to judge triple confidences. Moreover, the promising results of conventional translation-based methods on triple classification also confirm that positive triples should comply with the translation assumption well. Fig. \ref{Fig.sub.2.1} demonstrates the effective mechanism of local triple confidence. Since knowledge representations learned under our translation framework should follow the global consistency with all triples in KG, we can infer that \emph{William Shakespeare} is more likely to write \emph{Hamlet} rather than \emph{Pride and Prejudice}, even though there are noises in training set.

We assume that the more a triple fits the translation assumption, the more convincing this triple should be considered. To measure the local triple confidence during training, we first judge the current conformity of each triple with translation assumption. Inspired by the margin-based training strategy, we directly use the same pair-wise function to calculate the triple quality $Q(h,r,t)$ as follows:
\begin{equation}
\begin{split}
Q(h,r,t) = -(\gamma+E(h,r,t)-E(h',r',t')).
\end{split}
\end{equation}
A higher $Q(h,r,t)$ usually indicates a better triple judged by the translation assumption. At the beginning of training, we suppose all triples are correct, and initialize the local triple confidence $LT(h,r,t)$ for all triples as $1$. Since both entity and relation embeddings will change during training, the current local triple confidence for each triple should be also updated according to how much this triple fit the translation assumption. Formally, the local triple confidence $LT(h,r,t)$ changes with its triple quality $Q(h,r,t)$ as follows:
\begin{equation}
\begin{split}
LT(h,r,t) =
\begin{cases}
\alpha LT(h,r,t), \quad Q(h,r,t)\leqslant0.\\
LT(h,r,t)+\beta, \quad Q(h,r,t)>0.
\end{cases}
\end{split}
\end{equation}
$Q(h,r,t)\leqslant0$ implies the current triple doesn't comply with the translation rule, and thus the corresponding local triple confidence should decrease. On the contrary, $Q(h,r,t)>0$ encourages the triple to have a larger local triple confidence. Here, $\alpha \in (0,1)$ and $\beta > 0$ are hyper-parameters that control the ascend or descend pace of local triple confidence, with the assurance that $LT(h,r,t) \in (0,1]$. Note that the local triple confidence will decrease at a geometric rate while increase with a constant addition. It is because that we urge to punish the violations of translation rule, for those triples are more likely to be noises or conflicts, and thus should have smaller confidences.

\subsection{Global Path Confidence}

The local triple confidence is straightforward and effective, while simply concentrating on the inside of triples will fail to use rich global structural information in knowledge graphs. Moreover, local triple confidence won't work well with high noise rate. Therefore, we propose the global path confidence to take multi-step relation paths into consideration. For instance, in Fig. \ref{Fig.sub.2.2}, there are two multi-step relation paths from \emph{William Shakespeare} to \emph{Hamlet}. The lower path provides strong evidence to infer the relation \texttt{write}, while the upper path just provides weaker evidence. These paths could also help us to judge triple qualities.

A triple considered to have high global path confidence should follow two conditions: (1) it has more reliable paths from its head to tail entity, and (2) these reasoning paths are semantically closer to the corresponding relation. In the following sub-sections, we first introduce how to quantify the relation path reliability for each triple, and then propose two strategies using prior co-occurrence information and learned knowledge representations to measure the semantic similarity between paths and relations.

\subsubsection{Relation Path Reliability}

We assume that a relation path should be considered more important if it carries more information flow from the head to tail entity. Specifically, we follow the path-constraint resource allocation (PCRA) \cite{lin2015modeling} to measure the relation path reliability. The key idea of PCRA is inspired by resource allocation \cite{zhou2007bipartite}, which supposes there are certain resources associated with head entity $h$, and will flow throughout the whole knowledge graph via all relation paths. The resource amount that eventually flows to the tail entity $t$ via a certain path $p$ will be considered as the relation path reliability of $p$ given the entity pair $(h,t)$.

Formally, given a path $p=(r_1, \cdots, r_l)$ and entity pair $(h,t)$, the resource in $h$ will flow to $t$ through $l$ steps. Since there are probably multiple tails given head and relation, the path is represented as $E_0 \xrightarrow{r_1} \cdots \xrightarrow{r_l} E_l$, where $E_i$ represents the entity set at the $i$-th step, $E_0=\{h\}$ and $t \in E_l$. For entity $e \in E_i$, the resource $R_p(e)$ will be calculated as follows:
\begin{equation}
\begin{split}
R_p(e)=\sum_{e' \in E_{i-1}(\cdot,e)}\frac{R_p(e')}{|E_i(e',\cdot)|},
\end{split}
\end{equation}
in which $E_{i-1}(\cdot,e)$ represents the direct predecessors of $e$ via $r_i$, and $E_i(e',\cdot)$ represents the direct successors of $e'$ via $r_i$. All entities will be initialized with the same resource amount, and finally after $l$ steps from $h$ to $t$, the resource amount $R_p(t)$ is regarded as the relation path reliability $R(h,p,t)$ of $p$ with the given entity pair $(h,t)$.

\subsubsection{Prior Path Confidence}

We first introduce prior path confidence (PP), which utilize the co-occurrence of relation and path to represent their dissimilarity. We suppose that the more a relation occurs with a path, the more they are likely to represent similar semantic meanings. Formally, given a triple $(h,r,t)$ and its path set $S_{(h,t)}$ containing all paths between $h$ and $t$, the quality of the $i$-th relation-path pair $(r,p_i)$ is written as follows:
\begin{equation}
\begin{split}
Q_{PP}(r,p_i)=\epsilon+(1-\epsilon)\frac{P(r,p_i)}{P(p_i)},
\end{split}
\end{equation}
where $P(r,p_i)$ represents the prior probability of $r$ and $p_i$ co-occurrence, and $P(p_i)$ represents the prior probability of $p_i$ in KG. $\epsilon$ is a hyper-parameter for smoothing. Therefore, the prior path confidence (PP) is designed as follows:
\begin{equation}
\begin{split}
PP(h,r,t)=\sum_{p_i \in S_{(h,t)}}Q_{PP}(r,p_i) \cdot R(h,p_i,t).
\end{split}
\end{equation}
It indicates that the prior path confidence of $(h,r,t)$ depends on both relation-path similarities of all paths in $S_{(h,t)}$ and their corresponding relation path reliabilities. Note that since we merely consider the prior probabilities of paths and relations, prior path confidences are fixed during training.

\subsubsection{Adaptive Path Confidence}

The prior path confidence stays static during training, which is inflexible and may be strongly constrained by existing noises and conflicts in KGs. To address this problem, we propose the adaptive path confidence (AP) that could flexibly learn relation-path qualities according to their learned embeddings. Formally, given $r$ and $p_i=\{r_{i1}, \cdots, r_{ik}\}$, we directly represent the path embedding $\mathbf{p}_i$ with the sum of its relation embeddings $\mathbf{r}_{i1}+ \cdots + \mathbf{r}_{ik}$ under the translation assumption. The relation-path quality function of AP is defined as follows:
\begin{equation}
\begin{split}
Q_{AP}(r,p_i)=||\mathbf{r}-\mathbf{p}_i||=||\mathbf{r}-(\mathbf{r}_{i1}+ \cdots + \mathbf{r}_{ik})||.
\end{split}
\end{equation}
Since we assume that the relation embedding should be similar as the path embedding, a lower $Q_{AP}(r,p_i)$ implies a more convincing relation-path pair. The adaptive path confidence is then written as follows:
\begin{equation}
\begin{split}
AP(h,r,t)=\sigma(\sum_{p_i \in S_{(h,t)}} \frac{R(h,p_i,t)}{Q_{AP}(r,p_i)}),
\end{split}
\end{equation}
in which $\sigma(\cdot)$ stands for the sigmoid function. Adaptive path confidence can describe triple confidences dynamically with evolutionary relation embeddings during training, making our triple confidences more flexible and precise.

The overall triple confidence combines with all three kinds of confidences stated above. We have:
\begin{equation}
\begin{split}
C(h,r,t)=\lambda_1 \cdot LT(h,r,t) &+ \lambda_2 \cdot PP(h,r,t)\\
 &+ \lambda_3 \cdot AP(h,r,t),
\end{split}
\end{equation}
where $\lambda_1$, $\lambda_2$, $\lambda_3$ are hyper-parameters.

\subsection{Optimization and Implementation Details}

We utilize mini-batch stochastic gradient descent (SGD) to optimize our model. In training, all entity and relation embeddings could be either initialized randomly or pre-trained with TransE. For those entity pairs that don't have paths, we directly set their path-based confidences as $0$.

Path selection is essential in our model that will have significant impacts on the performances. Since the number of all paths grows exponentially with the increase of maximum path length, it is impractical to enumerate all paths in KG. Moreover, the path-based inference will be much weaker when the logical chain goes too far. Considering both effectiveness and efficiency, we limit the maximum length of paths to at most $2$-steps to prevent possible error propagation. Since relations in KGs are directed edges, we also consider those reverse relations when we detect relation paths.

\section{Experiment}


\subsection{Datasets}

In this paper, we evaluate our CKRL model based on FB15K \cite{bordes2013translating}, which is a typical benchmark knowledge graph extracted from Freebase \cite{bollacker2008freebase}. However, there are no explicit labelled noises or conflicts in FB15K. Therefore, we generate new datasets with different noise rates based on FB15K to simulate the real-world knowledge graphs constructed automatically with less human annotation.

Most noises and conflicts in real-world knowledge graphs derive from the misunderstanding between similar entities. It indicates that the noise (\emph{Jane Austen}, \texttt{write}, \emph{Hamlet}) is more likely to occur in real-world KG rather than (\emph{Soccer}, \texttt{write}, \emph{Hamlet}), in which the latter could be easily detected via entity type constraints. Inspired by the preprocessing in the evaluation task named triple classification, we construct negative triples (i.e., noises) following the same setting in \cite{socher2013reasoning}. Specifically, given a positive triple $(h,r,t)$ in KG, we randomly switch one of head or tail entities to form a negative triple $(h',r,t)$ or $(h,r,t')$. The generation of negative triples is constrained that $h'$ (or $t'$) should have appeared in the head (or tail) position with the same relation $r$ in dataset, which means that the head entity of relation \texttt{write} in negative triples should also be a writer. This constraint focuses on generating harder and more confusing cases, for those negative triples with mistype entities could be easily detected. We can directly utilize entity type information in Freebase or follow the local closed-world assumption \cite{krompass2015type} to collect type constraint information. Following this protocol, we construct three KGs based on FB15K with negative triples to be 10\%, 20\% and 40\% of positive triples, and then discard a small number of negative triples that violate type constraints. All three noisy datasets share the same entities, relations, validation and test sets with FB15K, with all generated negative triples fused into the original training set of FB15K. The statistics are listed in Table \ref{tab. 1}.

\begin{figure*}[!htbp]
\centering
\subfigure[FB15K-N1]{
\label{Fig.sub.1}
\includegraphics[width=0.32\textwidth]{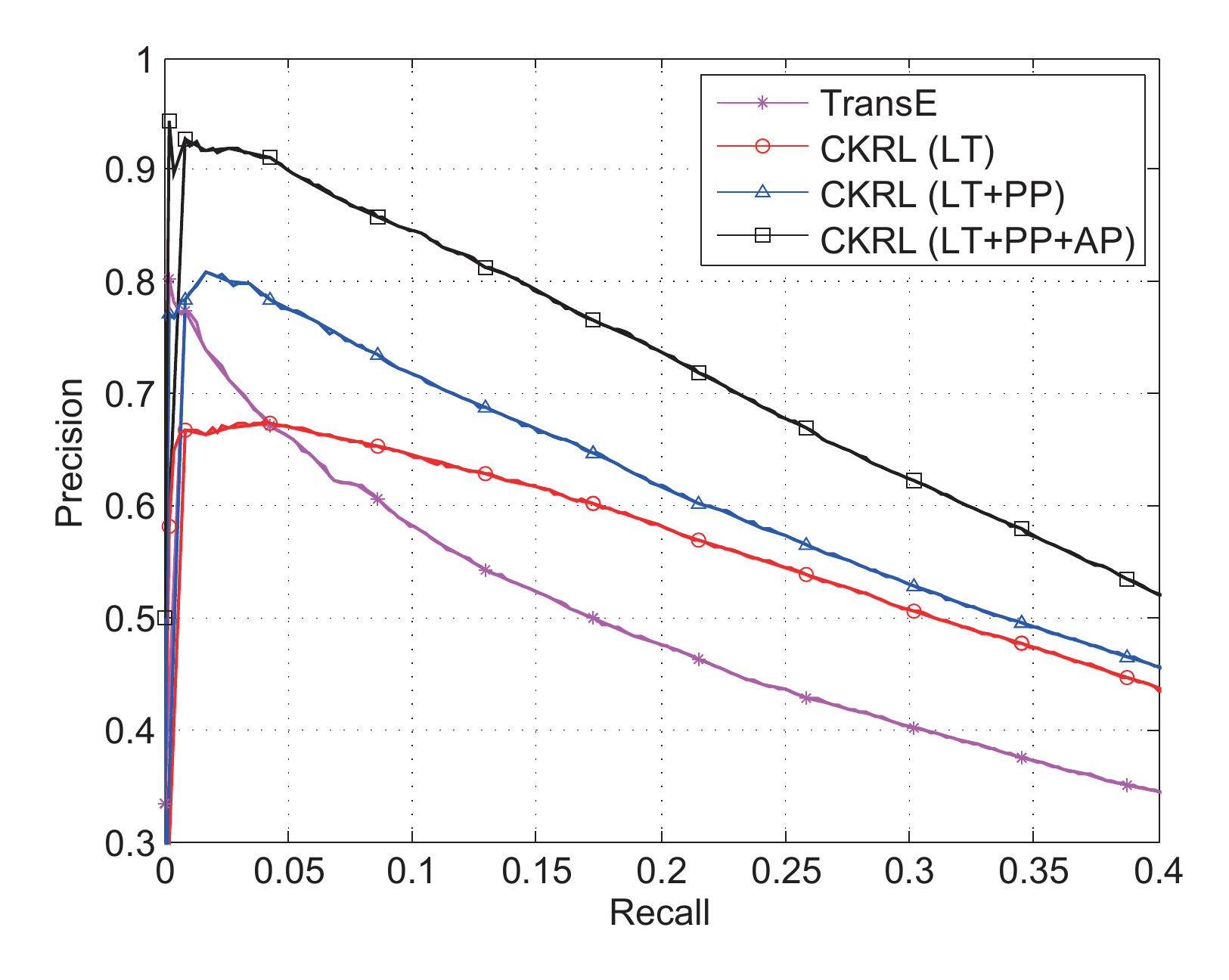}}
\subfigure[FB15K-N2]{
\label{Fig.sub.2}
\includegraphics[width=0.32\textwidth]{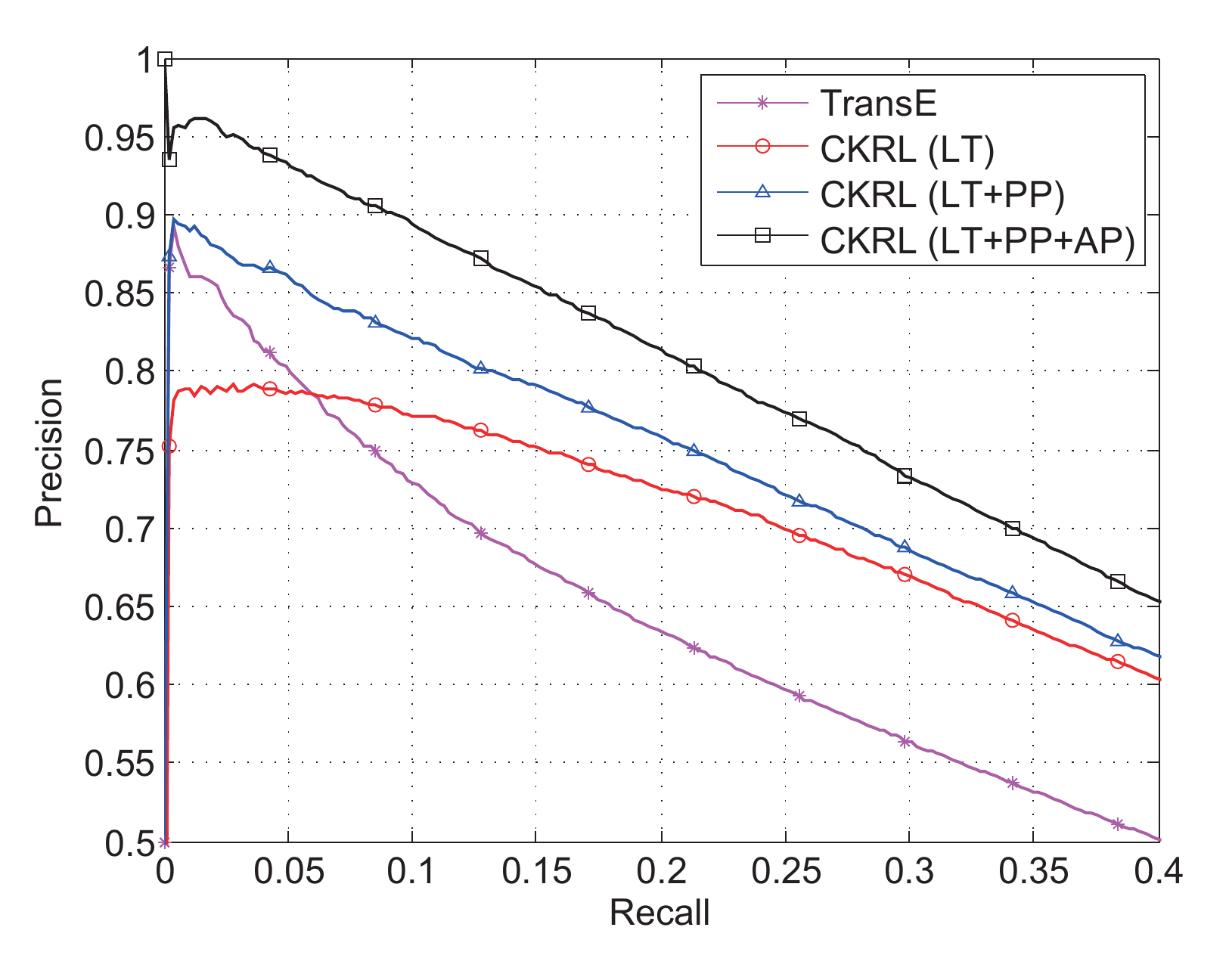}}
\subfigure[FB15K-N3]{
\label{Fig.sub.3}
\includegraphics[width=0.32\textwidth]{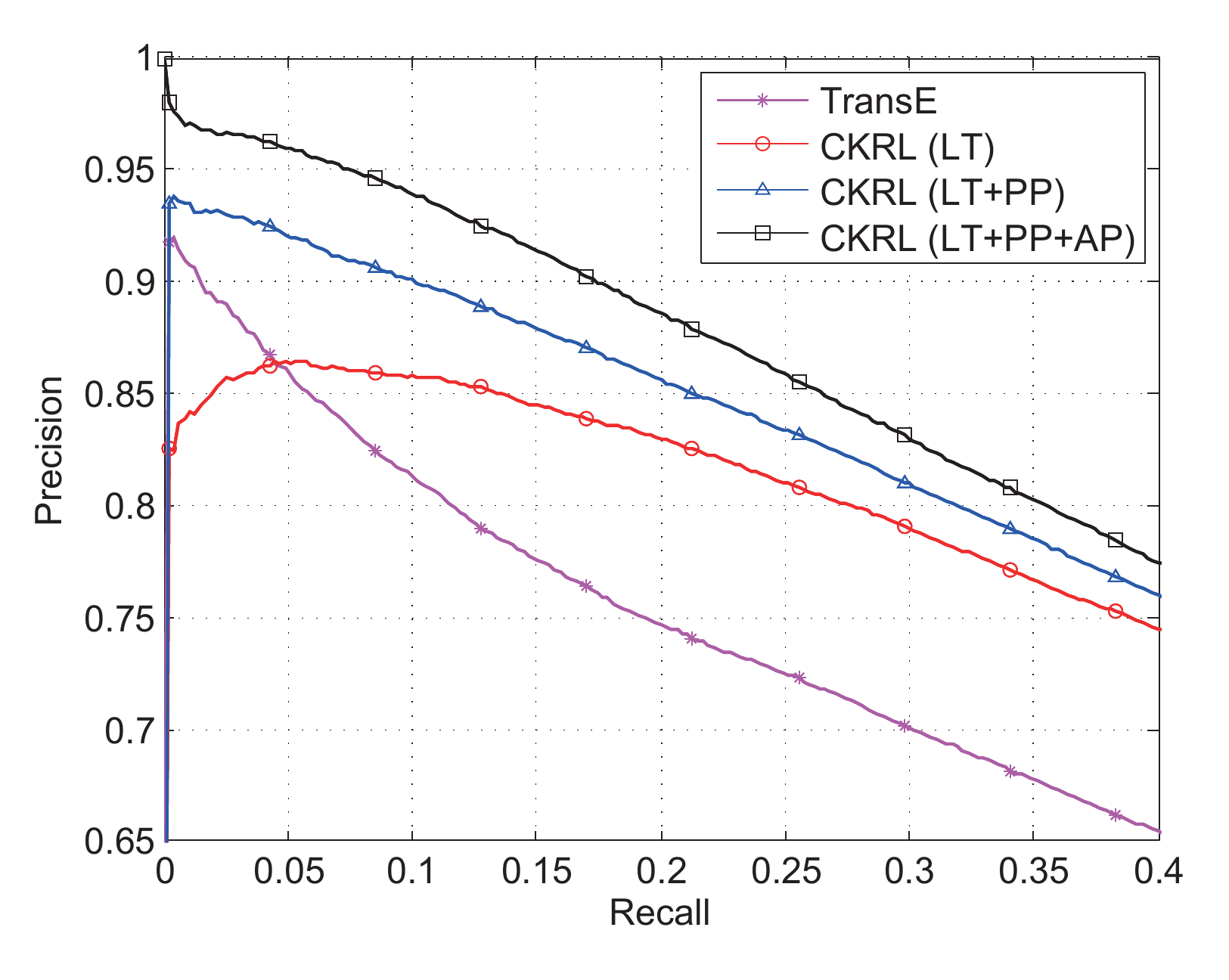}}
\caption{Evaluation results on knowledge graph noise detection.}
\label{fig. 3}
\end{figure*}

\begin{table}[!h]
\center
\small
\caption{\label{tab. 1} Statistics of datasets}

\begin{tabular}{cccccc}
 \toprule
  Dataset & \#Rel & \#Ent & \#Train & \#Valid & \#Test\\
 \midrule
  FB15K & 1,345 & 14,951 & 483,142 & 50,000 & 59,071\\
 \bottomrule
\end{tabular}

\begin{tabular}{p{48pt}<{\centering}p{42pt}<{\centering}p{42pt}<{\centering}p{42pt}<{\centering}}
 \toprule
  Datasets & FB15K-N1 & FB15K-N2 & FB15K-N3 \\
 \midrule
  \#Neg triple & 46,408 & 93,782 & 187,925 \\
 \bottomrule
\end{tabular}

\end{table}

\subsection{Experimental Settings}

In experiments, we evaluate our CKRL models with three different confidence combination strategies. CKRL (LT) represents the strategy which only considers local triple confidence, CKRL (LT+PP) considers both local triple confidence and prior path confidence, while CKRL (LT+PP+AP) considers all three kinds of triple confidences. We implement TransE \cite{bordes2013translating} as baseline for the CKRL learning framework is based on TransE, and it is not difficult for our confidence-aware framework to be utilized in other enhanced translation-based methods.

We train our CKRL model using mini-batch SGD with the margin $\gamma$ empirically set as $1.0$. We select the overall learning rate $\delta$ among $\{0.0005, 0.001, 0.002\}$, which is fixed during training. For local triple confidence, we select the descend controller $\alpha$ among $\{0.5, 0.7, 0.9\}$ and the ascend controller $\beta$ among $\{0.0001, 0.0005, 0.001\}$. For prior path confidence, the smoothing $\epsilon$ is empirically set as $0.01$. The optimal configurations of our models are: $\delta=0.001$, $\alpha=0.9$, $\beta=0.0001$, which are optimized on the validation set. We also evaluate various combination weights $\lambda_i$ when we calculate the overall triple confidence based on the three proposed methods. We select a unified weighting strategy for different evaluation tasks and datasets according to their overall performances to show the robustness of our CKRL models. Specifically, for CKRL (LT+PP), we select $\lambda_1=0.9$ and $\lambda_2=0.1$, while for CKRL (LT+PP+AP), we select $\lambda_1=1.5$, $\lambda_2=0.1$ and $\lambda_2=0.4$. For fair comparisons, the dimensions of both entity and relation embeddings in all models are equally set to be $50$.

\subsection{Knowledge Graph Noise Detection}

To verify the capability of our CKRL models in distinguishing noises and conflicts in knowledge graphs, we propose a novel evaluation task named knowledge graph noise detection. This task aims to detect possible noises in knowledge graphs according to their triple scores.

\subsubsection{Evaluation Protocol}

Inspired by the evaluation metric of triple classification in \cite{socher2013reasoning}, we consider the energy function scores $E(h,r,t)=||\mathbf{h}+\mathbf{r}-\mathbf{t}||$ as our triple scores, and then rank all triples in training set according to these scores. Those triples with higher scores are first considered to be noises. We utilize precision/recall curves to demonstrate the performances.

\subsubsection{Experimental Results}

Fig. \ref{fig. 3} demonstrates the results of knowledge graph noise detection, from which we can observe that: (1) our confidence-aware KRL models achieve the best performances on all three datasets with different noise proportions. It confirms the capability of our CKRL models in modeling triple confidence and detecting noises and conflicts in knowledge graphs. (2) CKRL (LT+PP+AP) has significant and consistent improvements in noise detection compared to other confidence-aware strategies. It indicates that the adaptive path confidence could provide more flexible and credible evidence for noise detection. Moreover, CKRL (LT+PP+AP) achieves impressively $84\sim94\%$ in precision with different noise proportions when the recall is $10\%$, which implies that our models could truly help in real-world KG noise detection. (3) CKRL (LT+PP) performs better than CKRL (LT) especially at the beginning of PR curves, which matters more in real-world KG noise detection systems. It implies that even though the local triple confidence is capable of capturing KG global consistency via learned knowledge representations, the global path information could still be a qualified supplement via multi-step path reasoning. (4) For further comparisons, we also evaluate PTransE \cite{lin2015modeling} which considers multi-step paths in KRL on this task with its energy function, while the results are surprisingly much worse than TransE. We find that PTransE can not detect noises in KGs well, for its path-based energy function scores only work when comparing positive and negative pairs. We don't show the results of PTransE due to the limited space.

\begin{table*}[!htbp]
\center
\small
\caption{\label{tab. 2} Evaluation results on entity prediction}
\begin{tabular}{c|cc|cc|cc|cc|cc|cc}
 \toprule
  Datasets & \multicolumn{4}{c|}{FB15K-N1} & \multicolumn{4}{c|}{FB15K-N2} & \multicolumn{4}{c}{FB15K-N3}\\
  \midrule
  \multirow{2}{*}{Metric} & \multicolumn{2}{c|}{Mean Rank} & \multicolumn{2}{c|}{Hits@10(\%)} & \multicolumn{2}{c|}{Mean Rank} & \multicolumn{2}{c|}{Hits@10(\%)} & \multicolumn{2}{c|}{Mean Rank} & \multicolumn{2}{c}{Hits@10(\%)}\\
   & Raw & Filter & Raw & Filter & Raw & Filter & Raw & Filter & Raw & Filter & Raw & Filter\\
  \midrule
  TransE & 240 & 144 & 44.9 & 59.8 & 250 & 155 & 42.8 & 56.3 & 265 & 171 & 40.2 & 51.8\\
  \midrule
  CKRL (LT) & 237 & 140 & \textbf{45.5} & \textbf{61.8} & 243 & 146 & \textbf{44.3} & \textbf{59.5} & \textbf{244} & \textbf{148} & 42.7 & \textbf{56.9}\\
  CKRL (LT+PP) & \textbf{236} & 139 & 45.3 & 61.6 & 241 & \textbf{144} & 44.2 & 59.4 & 245 & 149 & \textbf{42.8} & 56.8\\
  CKRL (LT+PP+AP) & \textbf{236} & \textbf{138} & 45.3 & 61.6 & \textbf{240} & \textbf{144} & 44.2 & 59.3 & 245 & 150 & \textbf{42.8} & 56.6\\
 \bottomrule
\end{tabular}
\end{table*}

\subsection{Knowledge Graph Completion}

Knowledge graph completion is a classical evaluation task that concentrates on the quality of knowledge representations \cite{bordes2012joint}. This task aims to complete a triple when one of head, tail or relation is missing, which can be viewed as a simple question answering task.

\subsubsection{Evaluation Protocol}

In this paper, we mainly focus on entity prediction, which is determined by the translation assumption that $\mathbf{h}+\mathbf{r}\simeq\mathbf{t}$. Following the same settings in \cite{bordes2013translating}, we conduct two measures as our evaluation metrics: (1) Mean Rank of correct entities, and (2) Hits@10 that indicates the proportion of correct answers ranked in top 10. We also follow the different evaluation settings of ``Raw" and ``Filter" utilized in \cite{bordes2013translating}.

\subsubsection{Experimental Results}

In Table \ref{tab. 2} we demonstrate the results of entity prediction with different noise rates, from which we can observe that: (1) all confidence-aware KRL models consistently and significantly outperform baseline on all noisy datasets with all evaluation metrics. It confirms the quality of learned knowledge representations, for they could not only detect noises in knowledge graphs, but also perform well in knowledge graph completion. (2) Comparing with evaluation results between different datasets, we find that the improvements introduced by our confidence-aware methods become more significant as the noise rate in KGs goes higher. It indicates that noises are harmful to entity prediction, and on the other hand reaffirms that considering triple confidence in knowledge representation learning is essential. (3) It seems that the global path confidence has few contributions on entity prediction. It may be partially caused by the uncertainty and incompleteness in path information due to possible error propagation and limited path selection. In parameter analysis, we find that a higher weight of global path confidence within a reasonable range will improve the performances of entity prediction while harming those of KG noise detection. Taking longer relation paths into consideration with prior knowledge or patterns to assure high path qualities will partially solve this problem. (4) For a more comprehensive comparison, we evaluate TransE on the original FB15K dataset (without any noises) with the same parameter settings, evaluation protocols and test set. The result of “Hits@10 (Filter)” is 64.7\% and “Mean Rank (Filter)” is 134, which can be viewed as the upper bound of our models. Compared to these results, the improvements of CKRL seem to be good as well.

\subsection{Triple Classification}

Triple classification aims to predict whether a triple in test set is correct or not according to the dissimilarity function, which could be viewed as a binary classification task. Triple classification could also be regarded as a simpler knowledge graph noise detection task in test set, for the noises in training set will influence the construction of knowledge representations, while the negative triples generated in test set will not.

\subsubsection{Evaluation Protocol}

Since there are no explicit negative triples in existing knowledge graphs, we construct negative triples in validation and test set following the same protocol in \cite{socher2013reasoning}. We also assure that the number of generated negative triples should be equal to that of positive triples. The classification is conducted as follows: we first learn different thresholds $\delta_r$ for each relation, which are optimized by maximizing the classification accuracies on validation set. In classification, if the energy function $||\mathbf{h}+\mathbf{r}-\mathbf{t}||<\delta_r$, the triple will be classified to be positive, and otherwise to be negative.

\begin{table}[!h]
\center
\small
\caption{\label{tab. 3} Evaluation results on triple classification}
\begin{tabular}{c|p{1.1cm}<{\centering}p{1.1cm}<{\centering}p{1.1cm}<{\centering}}
 \toprule
  Datasets & FB15K-N1 & FB15K-N2 & FB15K-N3 \\
  \midrule
  TransE & 81.3 & 79.4 & 76.9\\ 
  \midrule
  CKRL (LT) & 81.8 & \textbf{80.2} & 78.3\\ 
  CKRL (LT+PP) & \textbf{81.9} & 80.1 & \textbf{78.4}\\ 
  CKRL (LT+PP+AP) & 81.7 & \textbf{80.2} & 78.3\\ 
 \bottomrule
\end{tabular}
\end{table}

\subsubsection{Experimental Results}

Table \ref{tab. 3} demonstrates the results of triple classification. We can find that: (1) The CKRL models outperform baseline on all datasets, and the improvements become more significant with higher noise rates. It confirms that learning knowledge representations with triple confidence could also help for triple classification. (2) The advantages confidence-aware models have over baseline in this task seem to be smaller than those in KG noise detection. It is because that the CKRL models concentrate more on calculating confidences for triples in training set, but not for negative triples generated in test set. The possible improvements for CKRL in triple classification can only derive from better-learned knowledge representations, which are less straightforward compared to that in KG noise detection. Although CKRL models learn better knowledge representations, conventional models without confidence may also achieve comparable results.

\section{Conclusion and Future Work}

In this paper, we propose a novel CKRL model which aims to detect noises in knowledge graphs and learn robust knowledge representations simultaneously. To make our models more flexible and universal, we only consider the internal structural information in KGs to define the local triple confidence and the global path confidence. We evaluate our models on KG noise detection, KG completion and triple classification. Experimental results indicate that CKRL can well capture both local and global structural information to measure triple confidences, which is essential when detecting noises in KGs and learning better knowledge representations. The utilization of triple confidence could also inspire the noise detection in real-world knowledge construction. The source code and dataset of this paper can be obtained from \url{https://github.com/thunlp/CKRL}.

We will explore the following research directions in future: (1) external information such as entity attributes and entity descriptions could provide supplementary information to judge triple confidence. We will explore to combine external heterogeneous information with internal structural information to better understand entities and relations. (2) We observe that the experimental results on knowledge graph noise detection are promising even with high noise rate datasets. In future, we will extend our confidence-aware framework to noise detection in knowledge construction, which could protect knowledge graphs away from noises and conflicts via global structural information in KGs.


\section*{Acknowledgments}

This work is supported by the National Natural Science Foundation of China (NSFC No. 61572273, 61532010, 61661146007), Tsinghua University Initiative Scientific Research Program (20151080406), and the research fund of Tsinghua University - Tencent Joint Laboratory for Internet Innovation Technology. We also thank Yankai Lin, Lixin Zhang and the THUNLP team for their constructive advices.

\bibliographystyle{aaai}
\bibliography{reference}

\end{document}